# Localization of Networked Robot Systems Subject to Random Delay and Packet Loss

Manh Duong Phung, Thi Thanh Van Nguyen, Thuan Hoang Tran, and Quang Vinh Tran

*Abstract*— **This paper deals with the localization problem of mobile robot subject to communication delay and packet loss. The delay and loss may appear in a random fashion in both control inputs and observation measurements. A unified state-space representation is constructed to describe these mixed uncertainties. Based on it, the optimal linear estimator is developed. The main idea is the derivation of a relevance factor to incorporate delayed measurements to the being estimate. The estimator is then extended for nonlinear systems. The performance of this method is tested within the simulations in MATLAB and the experiments in a real robot system. The good localization results prove the efficiency of the method for the purpose of localization of networked mobile robot.**

## I. Introduction

Networked robot systems (NRSs) have gained the research interest recently due to its ability to support a wide range of applications including but not limited to telemedicine, telehomecare, virtual laboratory, and disaster rescue. A number of projects have been proposed to deal with the problems involved in the development of networked robots. Some works focus on hardware configuration and control architecture for specific applications [1]–[4]. The others deal with the navigation and transport protocols [5]–[8].

In this paper, the problem of localization is investigated. Differing from classic robots, localization of NRSs faces new challenges introducing by the network such as the inevitable communication delays, the out-of-sequence data arriving, the limited available bandwidth, and the partial intermittent observations. In the literature, several approaches have been proposed to cope with those changes.

In [9], four cameras are set in the robot field as external visual sensors to serve the localization and navigation task. The cameras are connected to the Internet and fixed on roof to formulate four adjacent grids of vision without the dead zone. A recognition algorithm is implemented to recognize and give out the relative location of the robot, target, landmark and obstacle symbols. In [10], the robot pose is estimated at the local site using odometry, sonar and compass sensors. The information is then transmitted to the remote site as the robot pose at the receiving time without considering the change of the robot during the communication time. A map-based localization method for Internet-based personal robot system is introduced in [11]. The absolute position of the robot is determined by comparing a reference map of the local site with the one built by a map building technique at the remote site.

It is recognizable from the proposed studies that the data transmission between the remote controller and the actuator was treated as a given condition rather than being modeled and analyzed to provide more theoretic approaches. In order to overcome this, several works have been proposed to address the communication problems such as the $H_\infty$ filter for systems with random delays, the optimal filter with multi packet dropout, or the adaptive Kalman filter with random sensor delays, multiple packet dropouts, and missing measurements [12]–[15]. Though these works are efficient for NRSs, the systems have to be linear. Further modifications are required to implement them to nonlinear systems such as mobile robot.

In this work, we develop a state estimator for the problem of robot localization subject to random delay and packet loss. An augmented filter called past observation-based extended Kalman filter (PO-EKF) inspired by the well-known optimal filter, the Kalman filter, is introduced as the compensator and estimator. The main idea behind the filter is the determination of a "relevance factor" which describes the relevance of observations from the past to the present. This factor is employed as a multiplier to enable the incorporation of delayed measurements to the *posteriori* estimation. Simulations have been carried out in MATLAB and experiments have been implemented in a real NRS. The results confirm the effectiveness of the proposed approach.

## II. System Modeling and Problem Formulation

Consider the following discrete time state-space robot system:

$$\mathbf{x}_{k+1} = f(\mathbf{x}_k, \mathbf{u}_k, \mathbf{w}_k) \qquad (1)$$

$$\mathbf{z}_k = h(\mathbf{x}_k, \mathbf{v}_k) \qquad (2)$$

where $\mathbf{x} = [x\, y\, \theta]^T$ is the state vector described the robot's pose (position and orientation), $\mathbf{z}_k$ is the measured output, $f$

M. D. Phung, T. T. V. Nguyen, T. H. Tran, and Q. V. Tran are with the Department of Electronics and Computer Engineering, VNU University of Engineering and Technology, Vietnam National University, Hanoi, Vietnam (Phone: +84437549272; email: duongpm@vnu.edu.vn).

and $h$ are the system functions, and $\mathbf{w}_k$ and $\mathbf{v}_k$ are independent, zero-mean, white noise processes with normal probability distributions: $\mathbf{w}_k \sim \mathbf{N}(0, Q_k)$; $\mathbf{v}_k \sim \mathbf{N}(0, R_k)$; $E(\mathbf{w}_i \mathbf{v}_j^T) = 0$.

When distributing over communication networks, the system is decentralized and its functioning operation depends on a number of network parameters such as delay, loss, and out of order. The networks are in general very complex and can greatly differ in their architecture and implementation. In this work, a network is modeled as a module between the plant and controller which delivers input signals and observation measurements with possible delay and loss. The delay is assumed to be random but measurable. The packet loss is modeled as a binary random variable $\lambda_k$ defined as follows:

$$\lambda_k = \begin{cases} 1, & \text{if a packet arrives at time } k \\ 0, & \text{otherwise} \end{cases} \qquad (3)$$

Let $n$ be the time delay (in number of sampling periods) between the controller and the actuator, $m$ be the time delay between the sensor and the controller, $\lambda_k^{ca}$ be the binary random variable described the arrival of inputs from the controller to the actuator, $\lambda_k^{sc}$ be the binary random variable described the arrival of measurements from the sensor to the controller. The NRS and its signal timing can be described as in Fig. 1.

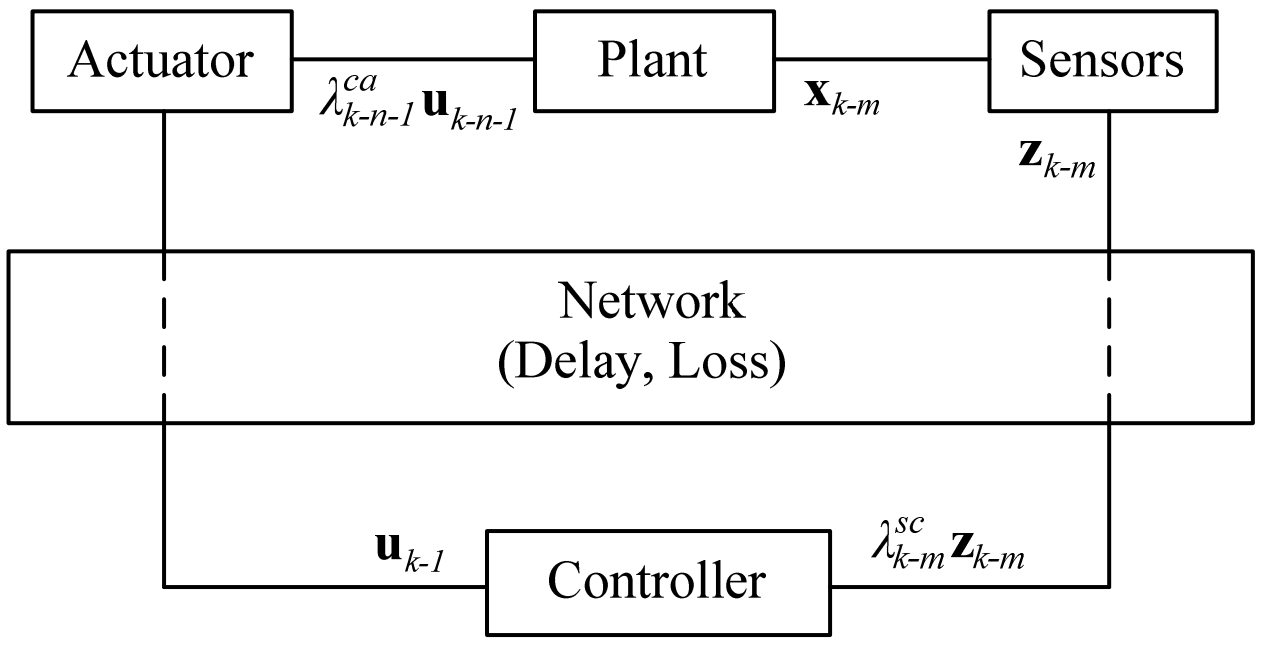

Fig. 1. Model of networked robot systems with signal timing

From Fig. 1, the plant at time $k$ is driven by the input at time $k$-$n$-$1$. The measurement received at the controller at time $k$ is taken at time $k$-$m$. Thus, the system is time-varying and can be rewritten as:

$$\mathbf{x}_k = f(\mathbf{x}_{k-1}, \lambda_{k-n-1}^{ca}\mathbf{u}_{k-n-1}, \mathbf{w}_{k-1}) \qquad (4)$$

$$\tilde{\mathbf{z}}_k = \lambda_{k-m}^{sc}\mathbf{z}_{k-m} = \lambda_{k-m}^{sc} h(\mathbf{x}_{k-m}, \mathbf{v}_{k-m}) \qquad (5)$$

The localization is the problem of state estimation of system (4)–(5). Our approach is the development of an optimal filter based on Kalman filter's theory [16].

## III. Localization with Past Observation-based Extended Kalman Filter

The Kalman filter is considered as one of the most efficient methods for mobile robot localization. It estimates the robot's state through two phases: prediction and correction. In each phase, it propagates an estimate value and an error covariance reflecting its belief to the estimate. When operating over the network, both inputs and observations suffer from the communication delay and loss. These data cannot be fused using the standard Kalman filter but require some modifications in the structure of the filter. In this section, we first derive the optimal filter for the linear NRS. The optimality is in sense that it minimizes the covariance of the estimation error. We then extend the filter for the nonlinear case.

### *A. Optimal filter for linear NRSs*

Consider linear NRSs. Equation (4)–(5) can be rewritten as:

$$\begin{aligned} \mathbf{x}_k &= A_{k-1}\mathbf{x}_{k-1} + \lambda_{k-n-1}^{ca} B_{k-1}\mathbf{u}_{k-n-1} + \mathbf{w}_{k-1} \\ &= A_{k-1}\mathbf{x}_{k-1} + B_{k-1}\tilde{\mathbf{u}}_{k-n-1} + \mathbf{w}_{k-1} \end{aligned} \qquad (6)$$

$$\begin{aligned} \tilde{\mathbf{z}}_k^i &= \lambda_{k-m}^{sc}\mathbf{z}_{k-m} = \lambda_{k-m}^{sc} H_{k-m}\mathbf{x}_{k-m} + \lambda_{k-m}^{sc}\mathbf{v}_{k-m} \\ &= \tilde{H}_i \mathbf{x}_i + \tilde{\mathbf{v}}_i \end{aligned} \qquad (7)$$

where $\tilde{\mathbf{u}}_k$, $\tilde{\mathbf{z}}_k^i$, $\tilde{H}_i$, $\tilde{\mathbf{v}}_i$, and $i$ are defined by the above equations. The optimal filter for system (6)–(7) is derived as follows.

**Priori State Estimate:** The *priori* estimate, $\hat{\mathbf{x}}_k^-$, is defined as the expectation of the state $\mathbf{x}_k$ given all measurements up to and including the last step *k-1*. From (6), it is given by:

$$\hat{\mathbf{x}}_k^- = E(\mathbf{x}_k) = A_{k-1}E(\mathbf{x}_{k-1}) + B_{k-1}E(\tilde{\mathbf{u}}_{k-n-1}) + E(\mathbf{w}_{k-1}) \qquad (8)$$

As $E(\mathbf{x}_{k-1})$ is the *posteriori* state estimate at time *k-1*, $\tilde{\mathbf{u}}_{k-n-1}$ is a known input, and $\mathbf{w}_{k-1}$ is zero-mean, (8) becomes:

$$\hat{\mathbf{x}}_k^- = A_{k-1}\hat{\mathbf{x}}_{k-1}^+ + B_{k-1}\tilde{\mathbf{u}}_{k-n-1} \qquad (9)$$

**Priori Error Covariance:** Let $\mathbf{e}_k^-$ and $\mathbf{e}_k^+$ be the *priori* and *posteriori* estimate errors, respectively:

$$\mathbf{e}_k^- = \mathbf{x}_k - \hat{\mathbf{x}}_k^- \qquad (10)$$

$$\mathbf{e}_k^+ = \mathbf{x}_k - \hat{\mathbf{x}}_k^+ \qquad (11)$$

From (6) and (9), the covariance of the *priori* estimate error is given by:

$$\begin{aligned} P_k^- &= E(\mathbf{e}_k^- \mathbf{e}_k^{-T}) \\ &= E(A_{k-1}\mathbf{e}_{k-1}^+ \mathbf{e}_{k-1}^{+T} A_{k-1}^T + \mathbf{w}_{k-1}\mathbf{w}_{k-1}^T \\ &\quad + A_{k-1}\mathbf{e}_{k-1}^+ \mathbf{w}_{k-1}^T + \mathbf{w}_{k-1}\mathbf{e}_{k-1}^{+T} A_{k-1}^T) \\ &= A_{k-1}P_{k-1}^+ A_{k-1}^T + Q_{k-1} \end{aligned} \qquad (12)$$

**Posteriori State Estimate:** From (7) the measurement $\tilde{\mathbf{z}}_k^i$ received at time $k$ would update the system state at a previous time $i$ rather than the present time $k$. Due to the delay, this measurement could not reach the estimator until time $k$. We therefore construct the data update equation of the form:

$$\hat{\mathbf{x}}_k^+ = \hat{\mathbf{x}}_k^- + K_k(\tilde{\mathbf{z}}_k^i - \tilde{H}_i \hat{\mathbf{x}}_i^-) \qquad (13)$$

and recompute the Kalman gain and error covariance to ensure the optimality of the new data update equation.

**Kalman gain and Posteriori Error Covariance**: Assume that the measurement is fused using (13) with an arbitrary gain $K_k$. The covariance of the *posteriori* estimate error, $P_k^+$, is determined as:

$$\begin{aligned} P_k^+ &= E(\mathbf{e}_k^+\mathbf{e}_k^{+T}) \\ &= E[\mathbf{e}_k^-\mathbf{e}_k^{-T} - \mathbf{e}_k^-\mathbf{e}_i^{-T}(K_k\tilde{H}_i)^T - \mathbf{e}_k^-\tilde{\mathbf{v}}_i^T K_k^T - K_k\tilde{H}_i\mathbf{e}_i^-\mathbf{e}_k^{-T} \\ &\quad + K_k\tilde{H}_i\mathbf{e}_i^-\mathbf{e}_i^{-T}(K_k\tilde{H}_i)^T + K_k\tilde{H}_i\mathbf{e}_i^-\tilde{\mathbf{v}}_i^T K_k^T - K_k\tilde{\mathbf{v}}_i\mathbf{e}_k^{-T} \\ &\quad + K_k\tilde{\mathbf{v}}_i\mathbf{e}_i^{-T}(K_k\tilde{H}_i)^T + K_k\tilde{\mathbf{v}}_i\tilde{\mathbf{v}}_i^T K_k^T] \end{aligned} \tag{14}$$

Due to the independence between $\mathbf{e}^-$ and $\tilde{\mathbf{v}}$, (14) can be simplified to:

$$\begin{aligned} P_k^+ &= E(\mathbf{e}_k^-\mathbf{e}_k^{-T}) - E(\mathbf{e}_k^-\mathbf{e}_i^{-T})(K_k\tilde{H}_i)^T - K_k\tilde{H}_i E(\mathbf{e}_i^-\mathbf{e}_k^{-T}) \\ &\quad + K_k\tilde{H}_i E(\mathbf{e}_i^-\mathbf{e}_i^{-T})(K_k\tilde{H}_i)^T + K_k E(\tilde{\mathbf{v}}_i\tilde{\mathbf{v}}_i^T)K_k^T] \\ &= P_k^- - L^T\tilde{H}_i^T K_k^T - K_k\tilde{H}_i L + K_k\tilde{H}_i P_i^-\tilde{H}_i^T K_k^T + K_k\tilde{R}_i K_k^T \end{aligned} \tag{15}$$

where $L = E(\mathbf{e}_i^-\mathbf{e}_k^{-T})$.

As the matrix $K_k$ is chosen to be the gain or blending factor that minimizes the *posteriori* error covariance, this minimization is accomplished by taking the derivative of the trace of the *posteriori* error covariance with respect to $K_k$, setting that result equal to zero, and then solving for $K_k$. Applying this process to (15) obtains:

$$\frac{\partial tr(P_k^+)}{\partial K_k} = 2(-L^T\tilde{H}_i^T + K_k\tilde{H}_i P_i\tilde{H}_i^T + K_k\tilde{R}_i) = 0 \tag{16}$$

$$\Leftrightarrow K_k = L^T\tilde{H}_i^T[\tilde{H}_i P_i^-\tilde{H}_i^T + \tilde{R}_i]^{-1} \tag{17}$$

Inserting (17) in (15) leads to a simpler form of $P_k^+$:

$$P_k^+ = P_k^- - K_k\tilde{H}_i L \tag{18}$$

In order to compute $L$, the *priori* state estimate at time $k$ needs determining from the estimate at time $i$. Through the time update (9) and the data update (13), $\mathbf{e}^-$ becomes:

$$\begin{aligned} \mathbf{e}_k^- &= \mathbf{x}_k - \hat{\mathbf{x}}_k^- \\ &= A_{k-1}\mathbf{e}_{k-1}^+ - \mathbf{w}_{k-1} \\ &= A_{k-1}[(I - K_{k-1}\tilde{H}_{k-1})\mathbf{e}_{k-1}^- + K_{k-1}\tilde{\mathbf{v}}_{k-1}] - \mathbf{w}_{k-1} \end{aligned} \tag{19}$$

After $m$ updating steps, the estimation error becomes:

$$\mathbf{e}_k^- = F\mathbf{e}_i^- + \xi_1(\mathbf{w}_i,...,\mathbf{w}_{k-1}) + \xi_2(\tilde{\mathbf{v}}_i,...,\tilde{\mathbf{v}}_{k-1}) \tag{20}$$

where

$$F = \prod_{j=1}^{m} A_{k-j}(I - K_{k-j}\tilde{H}_{k-j}) \tag{21}$$

and $\xi_1$ and $\xi_2$ are the functions of noise sequences $\mathbf{w}$ and $\tilde{\mathbf{v}}$. From (20) and the independence between $\mathbf{e}^-$ and noise sequences, the covariance $L$ becomes:

$$L = E(\mathbf{e}_i^-\mathbf{e}_k^{-T}) = P_i^- F^T \tag{22}$$

Substituting (22) in (18) and (17) yields:

$$P_k^+ = P_k^- - K_k\tilde{H}_i P_i^- F^T \tag{23}$$

and

$$K_k = FP_i^-\tilde{H}_i^T[\tilde{H}_i P_i^-\tilde{H}_i^T + \tilde{R}_i]^{-1} \tag{24}$$

- *Remark 1:* Equation (24) can be rewritten as:

  $$K_k = FK_i^* \tag{25}$$

  where $K_i^*$ is the Kalman gain at time $i$ of the standard Kalman filter. It turns out that the past residual $(\tilde{\mathbf{z}}_k - \tilde{H}_i\hat{\mathbf{x}}_i^-)$ in (13) can be normally updated to the *posteriori* estimate at time $k$ as at time $i$ but the Kalman gain needs to be changed by the factor $F$. This factor describes the relevant of the measurement updated at time $i$ to the state at time $k$.

- *Remark 2:* (13) can be rewritten as:

  $$\hat{\mathbf{x}}_k^+ = \hat{\mathbf{x}}_k^- + \lambda_i K_k(\mathbf{z}_k^i - H_i\hat{\mathbf{x}}_i^-) \tag{26}$$

  It implies that if a measurement is not arrived ($\lambda_i = 0$), the estimator does not use any information of the "dummy" observation to the estimate. It simply sets the *posteriori* estimate to the value of the *priori* estimate.

- *Remark 3:* When the delays $n$ and $m$ are zero, the new filter reduces to the standard form of the Kalman filter.

### *B. Optimal filter for nonlinear NRSs*

Though the filter derived in previous section is capable for NRSs, the system has to be linear. As practical robot system is often nonlinear, further modification needs to be accomplished. In this section, the derivation of the extended Kalman filter is inherited to extend the derived filter for nonlinear systems. The main idea is the linearization of a nonlinear system around its previous estimated states.

Performing a Taylor series expansion of the state equation around $(\hat{\mathbf{x}}_{k-1}^+, \tilde{\mathbf{u}}_{k-n-1}, 0)$ gives:

$$\begin{aligned} \mathbf{x}_k &= f(\hat{\mathbf{x}}_{k-1}^+, \tilde{\mathbf{u}}_{k-n-1}, 0) + \left.\frac{\partial f}{\partial \mathbf{x}}\right|_{(\hat{\mathbf{x}}_{k-1}^+, \mathbf{u}_{k-n-1}, 0)} (\mathbf{x}_{k-1} - \hat{\mathbf{x}}_{k-1}^+) \\ &\quad + \left.\frac{\partial f}{\partial \mathbf{w}}\right|_{(\hat{\mathbf{x}}_{k-1}^+, \mathbf{u}_{k-n-1}, 0)} \mathbf{w}_{k-1} \\ &= f(\hat{\mathbf{x}}_{k-1}^+, \tilde{\mathbf{u}}_{k-n-1}, 0) + A_{k-1}(\mathbf{x}_{k-1} - \hat{\mathbf{x}}_{k-1}^+) + W_{k-1}\mathbf{w}_{k-1} \\ &= A_{k-1}\mathbf{x}_{k-1} + [f(\hat{\mathbf{x}}_{k-1}^+, \tilde{\mathbf{u}}_{k-n-1}, 0) - A_{k-1}\hat{\mathbf{x}}_{k-1}^+] + W_{k-1}\mathbf{w}_{k-1} \\ &= A_{k-1}\mathbf{x}_{k-1} + \tilde{\mathbf{u}}_{k-n-1}^* + \mathbf{w}_{k-1}^* \end{aligned} \tag{27}$$

where $A_{k-1}, W_{k-1}$ are defined by the above equation. Similarly, the measurement equation is linearized around $(\hat{\mathbf{x}}_i^-, 0)$ to obtain

$$\tilde{\mathbf{z}}_k^i = \lambda_i[h(\hat{\mathbf{x}}_i^-,0) + \frac{\partial h}{\partial \mathbf{x}}\Big|_{(\hat{\mathbf{x}}_i^-,0)}(\mathbf{x}_i - \hat{\mathbf{x}}_i^-) + \frac{\partial h}{\partial \mathbf{v}}\Big|_{(\hat{\mathbf{x}}_i^-,0)}\mathbf{v}_i]$$
$$= \tilde{h}(\hat{\mathbf{x}}_i^-,0) + \tilde{H}_i(\mathbf{x}_i - \hat{\mathbf{x}}_i^-) + \tilde{V}_i\mathbf{v}_i \quad (28)$$
$$= \tilde{H}_i\mathbf{x}_i + [\tilde{h}(\hat{\mathbf{x}}_i^-,0) - \tilde{H}_i\hat{\mathbf{x}}_i^-] + \tilde{V}_i\mathbf{v}_i$$
$$= \tilde{H}_i\mathbf{x}_i + \boldsymbol{\varepsilon}_i^* + \tilde{\mathbf{v}}_i^*$$

where $\tilde{h}_i$, $\tilde{H}_i$, $\tilde{V}_i$ are defined by the above equation. The system (27) and the measurement (28) now become linear and the proposed filter can be applied to obtain the new filter for nonlinear NRSs as follows:

The time update equations at prediction phase:

$$\hat{\mathbf{x}}_k^- = f(\hat{\mathbf{x}}_{k-1}^+, \tilde{\mathbf{u}}_{k-n-1}, \mathbf{0}) \quad (29)$$

$$P_k^- = A_{k-1}P_{k-1}^+A_{k-1}^T + W_{k-1}Q_{k-1}W_{k-1}^T \quad (30)$$

The data update equations at correction phase:

$$F = \prod_{j=1}^{m} A_{k-j}(I - K_{k-j}\tilde{H}_{k-j}) \quad (31)$$

$$K_k = FP_i^-\tilde{H}_i^T(\tilde{H}_iP_i^-\tilde{H}_i^T + \tilde{V}_i\tilde{R}_i\tilde{V}_i^T)^{-1} \quad (32)$$

$$\hat{\mathbf{x}}_k^+ = \hat{\mathbf{x}}_k^- + K_k[\tilde{\mathbf{z}}_k^i - \tilde{h}(\hat{\mathbf{x}}_i^-, \mathbf{0})] \quad (33)$$

$$P_k^+ = P_k^- - K_k\tilde{H}_iP_i^-F^T \quad (34)$$

We call this filter past observation-based extended Kalman filter (PO-EKF).

## IV. Simulations

In order to evaluate the efficiency of the PO-EKF for the localization of NRSs, simulations have been carried out in MATLAB.

### A. Simulation Setup

Parameters for simulations are extracted from the real NRS described in the next section. The robot is a two wheeled, differential-drive mobile robot with the kinematics given by:

$$x_{k+1} = x_k + \frac{R}{2}T_s(\omega_L(k) + \omega_R(k))\cos\theta_k$$
$$y_{k+1} = y_k + \frac{R}{2}T_s(\omega_L(k) + \omega_R(k))\sin\theta_k \quad (35)$$
$$\theta_{k+1} = \theta_k + \frac{R}{L}T_s(\omega_L(k) - \omega_R(k))$$

where $T_s$ is the sampling period, $R$ is the radius of driven wheels, $L$ is the distance between the wheels, and $\omega_L(k)$ and $\omega_R(k)$ are the rotational speeds of the left and right wheels, respectively. The input noise is modeled as being proportional to the angular speed $\omega_L(k)$ and $\omega_R(k)$ of the left and right wheels, respectively. The covariance matrix $Q_k$ is defined as:

$$Q_k = \begin{bmatrix} \delta\omega_R^2(k) & 0 \\ 0 & \delta\omega_L^2(k) \end{bmatrix} \quad (36)$$

where $\delta$ is a constant with the value 0.01 determined by experiments. In the simulations, it is supposed that the robot has a sensor system that can measure the robot position and orientation in the motion plane. The measurements suffer from a Gaussian noise with zero mean and the covariance:

$$R_k = \begin{bmatrix} 0.01 & 0 & 0 \\ 0 & 0.01 & 0 \\ 0 & 0 & 0.018 \end{bmatrix} \quad (37)$$

Remaining parameters are retrieved from the kinematics of the robot as follows:

$$A_{k+1} = \frac{\partial f_k}{\partial \mathbf{x}}\Big|_{(\hat{\mathbf{x}}_k^+,\mathbf{u}_k,\mathbf{0})} = \begin{bmatrix} 1 & 0 & -T_s v_c \sin\hat{\theta}_k^+ \\ 0 & 1 & T_s v_c \cos\hat{\theta}_k^+ \\ 0 & 0 & 1 \end{bmatrix} \quad (38)$$

$$W_{k+1} = \frac{\partial f_k}{\partial \mathbf{w}}\Big|_{(\hat{\mathbf{x}}_k^+,\mathbf{u}_k,\mathbf{0})} = T_s\frac{R}{2}\begin{bmatrix} \cos\hat{\theta}_k^+ & \cos\hat{\theta}_k^+ \\ \sin\hat{\theta}_k^+ & \sin\hat{\theta}_k^+ \\ \frac{2}{L} & \frac{2}{L} \end{bmatrix} \quad (39)$$

$$H_k = V_k = I \quad (40)$$

In simulations, the performance of the PO-EKF is compared with the EKF and the optimal filter proposed in [15]. The EKF is implemented with the assumption that it does not know if a measurement is delayed or not. It incorporates every received measurement as there is no delay. The optimal filter in [15] is called LEKF in this work.

### B. Simulation Results

The first simulation is conducted with network parameters as follows: the time delay is between 100ms and 800ms, and the loss rate is 1%. These values are determined based on experimental measurements of the Internet (as will be described in next section). Fig. 2a gives the tracking performance of the filters in motion plane. Fig. 2b, c, d give the comparison curves of the root mean square errors (RMSEs) simulated by 100-times Monte Carlo tests for filters. We see that EKF has the worst accuracy while the PO-EKF and LEKF introduce similar property. Table I shows the amount of floating point operations and the execution time of the filters, scaled with respect to the EKF. The PO-EKF is around two times higher than the EKF but hundred times smaller than the LEKF.

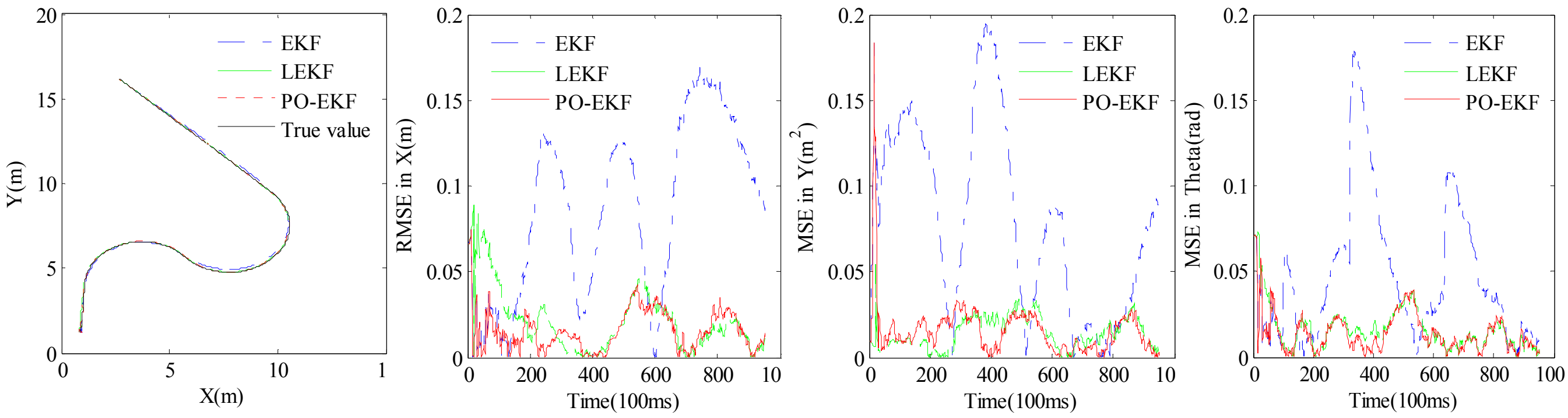

Fig. 2. Comparison between our filter (PO-EKF) and the EKF and the LEKF with $m = n = [1, 8]$, $p(\lambda^{ca} = 0) = p(\lambda^{sc} = 0) = 0.01$
(a) Trajectories in motion plan; (b) RMSE in X direction; (c) RMSE in Y direction; (d) RMSE in orientation

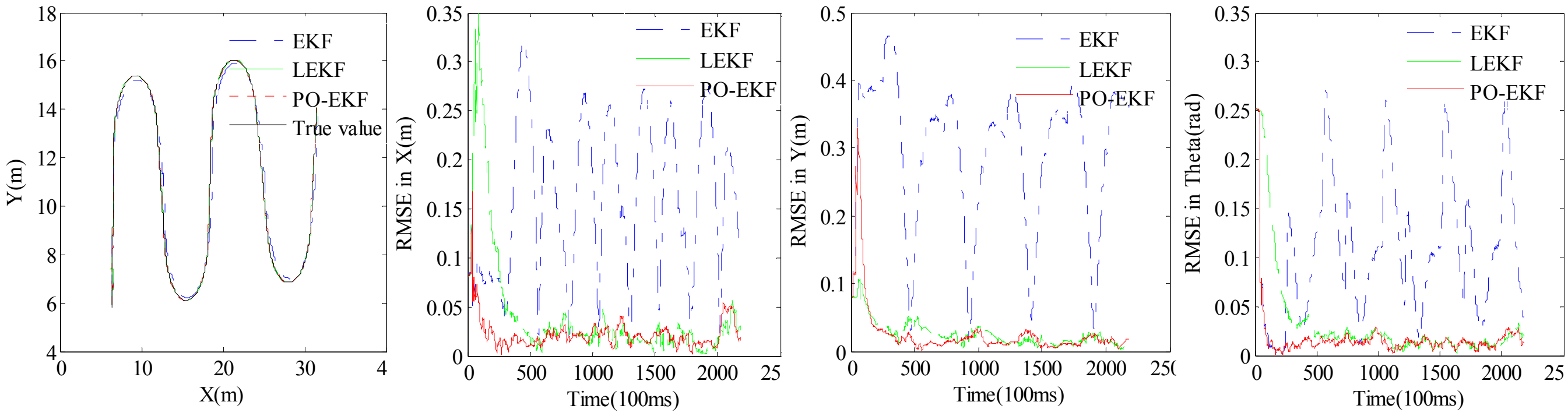

Fig. 3. Comparison between our filter (PO-EKF) and the EKF and the LEKF with $m = n = [8, 15]$, $p(\lambda^{ca} = 0) = p(\lambda^{sc} = 0) = 0.1$
(a) Trajectories in motion plan; (b) RMSEs in X direction; (c) RMSEs in Y direction; (d) RMSEs in orientation

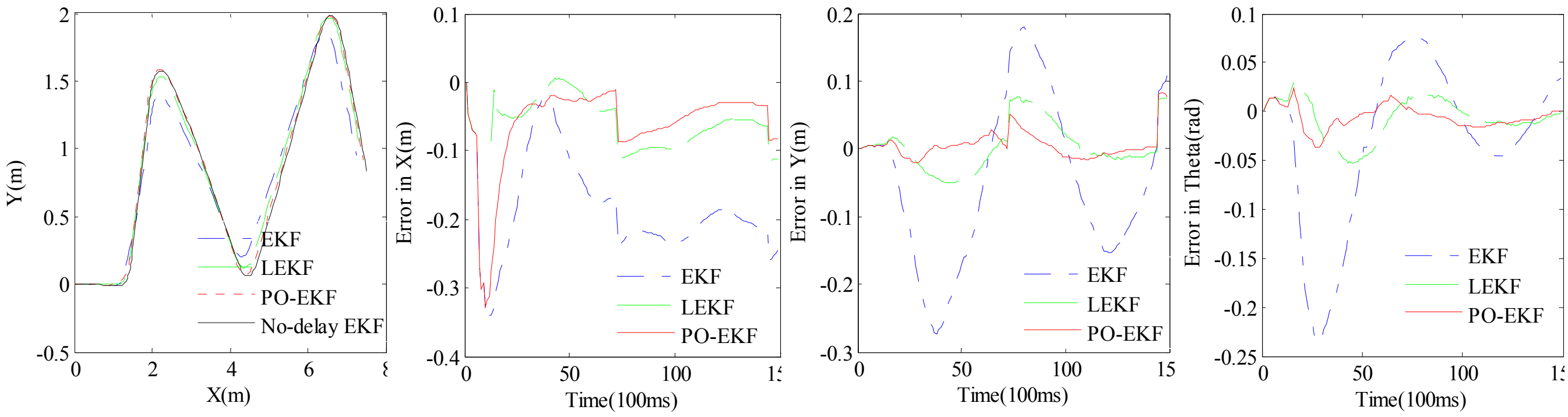

Fig. 4. Comparison between our filter (PO-EKF) and the EKF and the LEKF with local configuration: $m = n = [3, 5]$, $p(\lambda^{ca} = 0) = p(\lambda^{sc} = 0) = 0.015$
(a) Trajectories in motion plan; (b) Errors in X direction; (c) Errors in Y direction; (d) Errors in orientation

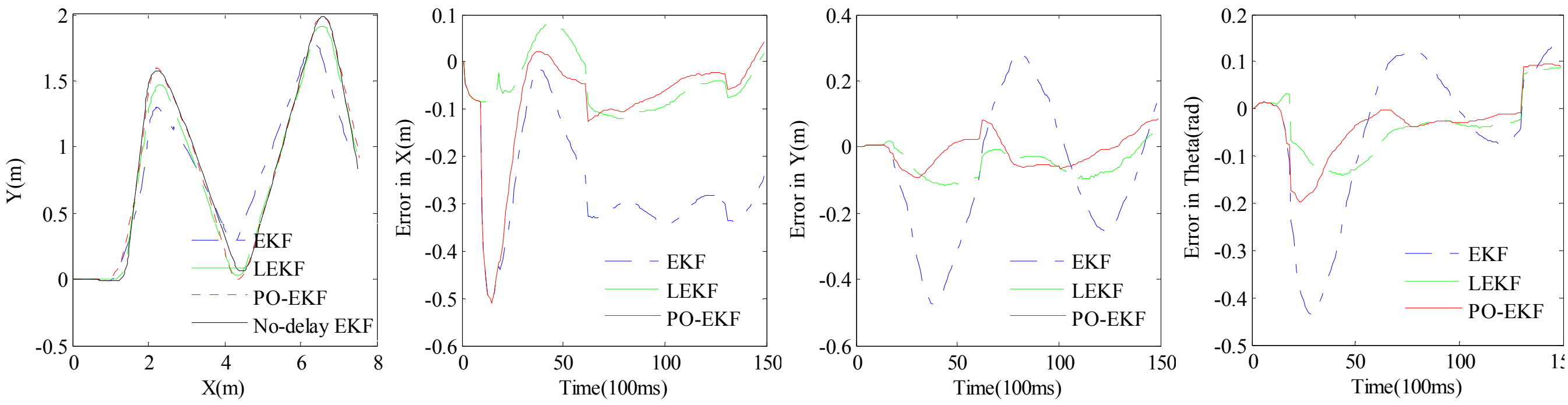

Fig. 5. Comparison between our filter (PO-EKF) and the EKF and the LEKF with VPN configuration: $m = n = [6, 8]$, $p(\lambda^{ca} = 0) = p(\lambda^{sc} = 0) = 0.02$
(a) Trajectories in motion plan; (b) Errors in X direction; (c) Errors in Y direction; (d) Errors in orientation

TABLE I. NORMALIZED COMPUTATIONAL BURDEN OF FILTERS

| | EKF | LEKF | PO-EKF |
|---|---|---|---|
| Floating point operations | 1.0 | 478.7 | 2.4 |
| Execution time | 1.0 | 532.3 | 2.0 |

In the next simulation, an extreme scenario is considered in which the time delay is between 800ms and 1500ms and the loss rate is 10%. The robot follows a sinusoidal path. The LEKF uses a finite buffer with 50 slots. Fig. 2 shows the tracking performance and the RMSEs. Table II shows the computational burden of filters. We see that the PO-EKF has better accuracy than the EKF and the same accuracy as the LEKF at steady state. Though the LEKF reduces the computation (based on the finite buffer), it is still high compared to the PO-EKF.

TABLE II. NORMALIZED COMPUTATIONAL BURDEN OF FILTERS

| | EKF | LEKF | PO-EKF |
|---|---|---|---|
| Floating point operations | 1.0 | 36.5 | 4.7 |
| Execution time | 1.0 | 33.7 | 2.4 |

## V. EXPERIMENTS

Experiments have been carried out in a real NRS. Details of the NRS can be referred from our previous work [17]. Two network configurations were employed in experiments. One is the local configuration in which the robot and the controller are connected to local Internet service providers. The other is the VPN configuration in which the robot and the controller are connected (via VPNs) to servers located at the United State. The purpose is to capture the low and high delay of the network. Due to the fact that every attempt to measure the true trajectory in experiments is influenced by measurement errors, trajectories estimated by filters are compared with the trajectory estimated by the EKF with no-delay data.

Fig. 4–5 show the localization results in local and VPN configurations, respectively. We see that the PO-EKF introduces better accuracy than the EKF and the same accuracy as the LEKF at steady state.

## VI. CONCLUSION

In this paper, we introduce a new state estimator called PO-EKF for the localization of NRSs subject to random delay and packet loss. The optimality of the estimator was theoretically proven. The good performance in term of accuracy and computational demand was confirmed through a number of simulations, comparisons, and experiments.

## ACKNOWLEDGMENT

This work was technically supported by the project CN.12.15 of VNU University of Engineering and Technology. The travel grant was supported by the Foundation for Science and Technology Development (NAFOSTED).